\documentclass[10pt,twocolumn,letterpaper]{article}

\usepackage{cvpr}
\usepackage{times}
\usepackage{epsfig}
\usepackage{graphicx}
\usepackage{amsmath}
\usepackage{amssymb}
\usepackage{multirow}
\usepackage{makecell}

\usepackage[breaklinks=true,bookmarks=false]{hyperref}

\cvprfinalcopy % *** Uncomment this line for the final submission

\def\cvprPaperID{****} % *** Enter the CVPR Paper ID here

\begin{document}

%%%%%%%%% TITLE
\title{Bi-Directional Cascade Network for Perceptual Edge Detection}

\author{Jianzhong He\textsuperscript{1},Shiliang Zhang\textsuperscript{1},Ming Yang\textsuperscript{2},Yanhu Shan\textsuperscript{2},Tiejun Huang\textsuperscript{1}\\
\textsuperscript{1}Peking University, \textsuperscript{2}Horizon Robotics, Inc.\\
{\tt\small {\{jianzhonghe,slzhang.jdl,tjhuang\}}@pku.edu.cn,m-yang4@u.northwestern.edu, yanhu.shan@gmail.com}
}

\maketitle
\def\eg{\emph{e.g.}}
\def\ie{\emph{i.e.}}
\def\etc{\emph{etc.}}
\def\etal{\emph{et~al.}}
\def\L{\mathcal{L}}
%%%%%%%%% ABSTRACT
\begin{abstract}
%the two componet whether reverse 1)BDCN 2)SEM
Exploiting multi-scale representations is critical to improve edge detection for objects at different scales. To extract edges at dramatically different scales, we propose a Bi-Directional Cascade Network (BDCN) structure, where an individual layer is supervised by labeled edges at its specific scale, rather than directly applying the same supervision to all CNN outputs. Furthermore, to enrich multi-scale representations learned by BDCN, we introduce a Scale Enhancement Module (SEM) which utilizes dilated convolution to generate multi-scale features, instead of using deeper CNNs or explicitly fusing multi-scale edge maps. These new approaches encourage the learning of multi-scale representations in different layers and detect edges that are well delineated by their scales. Learning scale dedicated layers also results in compact network with a fraction of parameters. We evaluate our method on three datasets, i.e., \emph{BSDS500}, \emph{NYUDv2}, and \emph{Multicue}, and achieve ODS F-measure of 0.828, 1.3\% higher than current state-of-the art on BSDS500. The code has been available\footnote{https://github.com/pkuCactus/BDCN.}.
\end{abstract}

%-------------------------------------------------------------------------
\section{Introduction}

\label{sec:intro}
Edge detection targets on extracting object boundaries and perceptually salient edges from natural images, which preserve the gist of an image and ignore unintended details. Thus, it is important to a variety of mid- and high-level vision tasks, such as image segmentation~\cite{arbelaez2011contour,rother2004grabcut}, object detection and recognition~\cite{ferrari2008groups,girshick2014rich}, \emph{etc}. Thanks to research efforts ranging from exploiting low-level visual cues with hand-crafted features~\cite{canny1986computational,kittler1983accuracy,arbelaez2011contour,lim2013sketch,dollar2015fast} to recent deep models~\cite{bertasius2015high,Liu_2017_CVPR,kokkinos2015pushing,Wang_2017_CVPR}, the accuracy of edge detection has been significantly boosted. For example, on the Berkeley Segmentation Data Set and Benchmarks 500 (\emph{BSDS500})~\cite{arbelaez2011contour}, the detection performance has been boosted from 0.598~\cite{comaniciu2002mean} to 0.815~\cite{Wang_2017_CVPR} in ODS F-measure.
%Some recent methods \cite{kokkinos2015pushing,Liu_2017_CVPR,Wang_2017_CVPR} have outperformed the human accuracy (ODS F-measure of 0.803  \cite{arbelaez2011contour}) on \emph{BSDS}.

\begin{figure}[t]
\centering
\centerline{\epsfig{figure=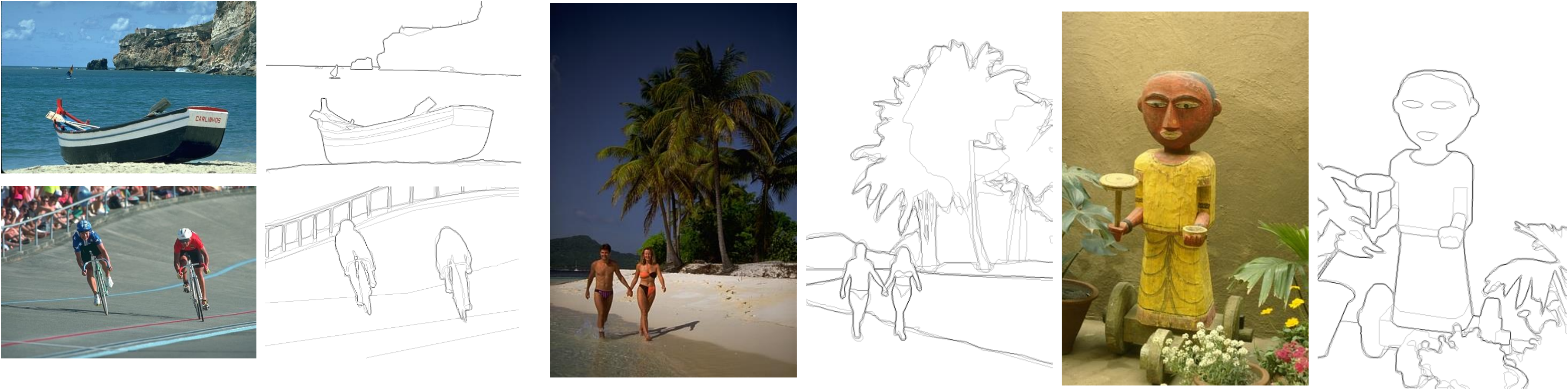, width=8.3cm}}
\caption{Some images and their ground truth edge maps in \emph{BSDS500} dataset. The scale of edges in one image varies considerably, like the boundaries of human body and hands.}
\label{fig:fig_vis}
\end{figure}
Nevertheless, there remain some open issues worthy of studying. As shown in Fig.~\ref{fig:fig_vis}, edges in one image stem from both object-level boundaries and meaningful local details, \emph{e.g.}, the silhouette of human body and the shape of hand gestures. The variety of scale of edges makes it crucial to exploit multi-scale representations for edge detection. Recent neural net based methods~\cite{Bertasius2015DeepEdge,shen2015deepcontour,xie2015holistically} utilize hierarchal features learned by Convolutional Neural Networks (CNN) to obtain multi-scale representations. To generate more powerful multi-scale representation, some researchers adopt very deep networks, like ResNet50~\cite{he2016deep}, as the backbone model of the edge detector. Deeper models generally involve more parameters, making the network hard to train and expensive to infer. Another way is to build an image pyramid and fuse multi-level features, which may involve redundant computations. In another word, can we use a shallow or light network to achieve a comparable or even better performance?
%It is appealing to explore an efficient multi-scale representation for edge detection using end-to-end CNN learning.

Another issue is about the CNN training strategy for edge detection, \emph{i.e.}, supervising predictions of different network layers by one general ground truth edge map~\cite{xie2015holistically,Liu_2017_CVPR}. For instance, HED~\cite{xie2015holistically,xie2017ijcv} and RCF~\cite{Liu_2017_CVPR} compute edge prediction on each intermediate CNN output to spot edges at different scales, \emph{i.e.}, the lower layers are expected to detect more local image patterns while higher layers capture object-level information with larger receptive fields. Since different network layers attend to depict patterns at different scales, it is not optimal to train those layers with the same supervision. In another word, existing works~\cite{xie2015holistically,xie2017ijcv,Liu_2017_CVPR} enforce each layer of CNN to predict edges at all scales and ignore that one specific intermeadiate layer can only focus on edges at certain scales.
Liu \etal~\cite{liu2016learning} propose to relax the supervisions on intermediate layers using Canny~\cite{canny1986computational} detectors with layer-specific scales. However, it is hard to decide layer-specific scales through human intervention.
%Liu \etal~\cite{liu2016learning} propose to relax the supervision for intermediate layers with labeled edges at manually selected scales. However, it is hard to select optimal scales through human intervention.

Aiming to fully exploit the multiple scale cues with a shallow CNN, we introduce a Scale Enhancement Module (SEM) which consists of multiple parallel convolutions with different dilation rates. As shown in image segmentation~\cite{chen2016deeplab}, dilated convolution effectively increases the size of receptive fields of network neurons. By involving multiple dilated convolutions, SEM captures multi-scale spatial contexts. Compared with previous strategies, \ie, introducing deeper networks and explicitly fusing multiple edge detections, SEM does not significantly increase network parameters and avoids the repetitive edge detection on image pyramids.
%Of course, to further boost performance, we can still employ the image pyramid idea and sacrifice inference time.

\begin{figure}[t]
\centering
\centerline{\epsfig{figure=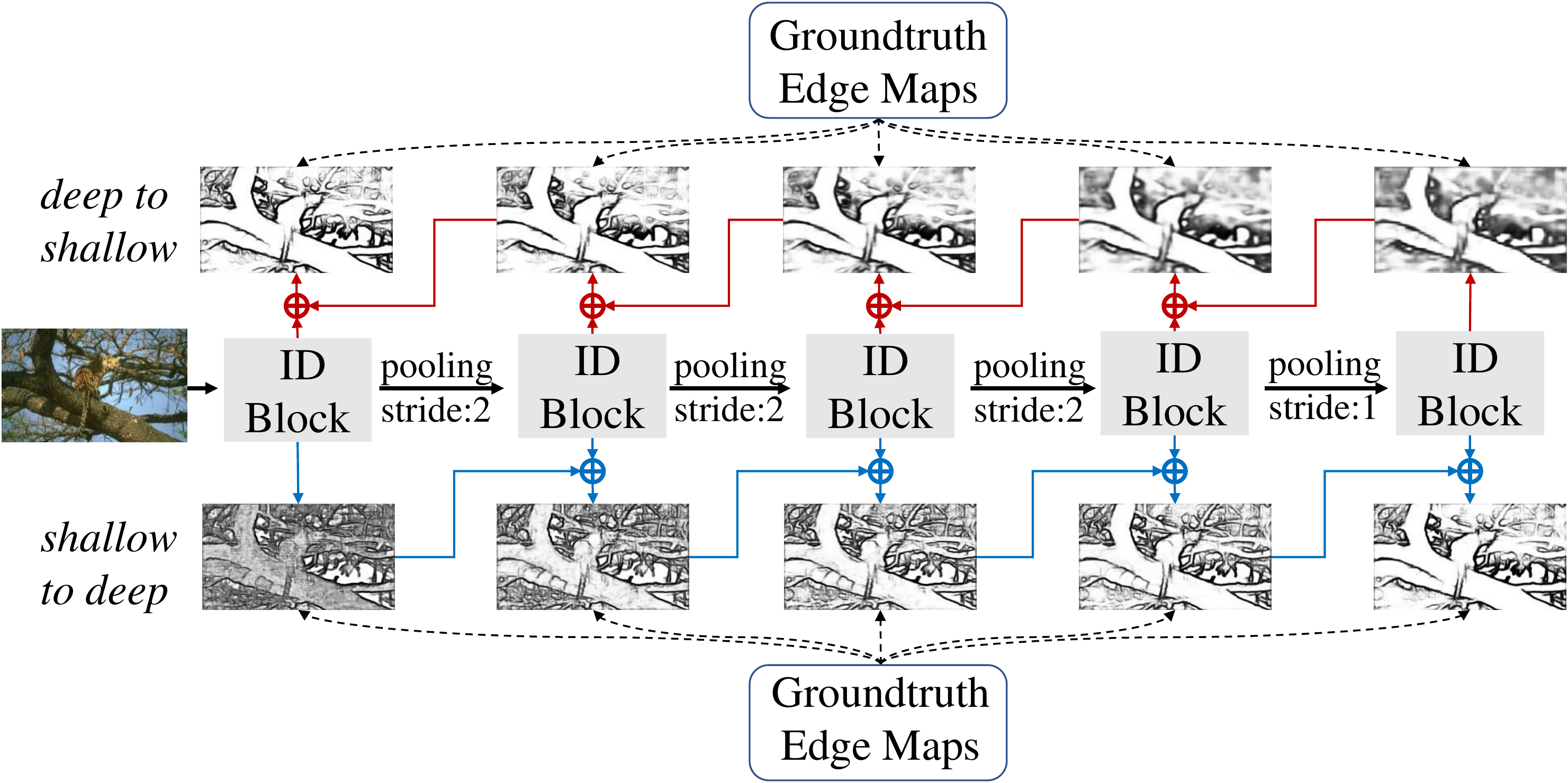, width=8.3cm}}
\caption{The overall architecture of BDCN. ID Block denotes the Incremental Detection Block, which is the basic component of BDCN. Each ID Block is trained by layer-specific supervisions inferred by a bi-directional cascade structure. This structure trains each ID Block to spot edges at a proper scale. The predictions of ID Blocks are fused as the final result.}
\label{fig:fig_overall}
\end{figure}

To address the second issue, each layer in CNN shall be trained by proper layer-specific supervision, \emph{e.g.}, the shallow layers are trained to focus on meaningful details and deep layers should depict object-level boundaries. We propose a Bi-Directional Cascade Network (BDCN) architecture to achieve effective layer-specific edge learning. For each layer in BDCN, its layer-specific supervision is inferred by a bi-directional cascade structure, which propagates the outputs from its adjacent higher and lower layers, as shown in Fig.~\ref{fig:fig_overall}. In another word, each layer in BDCN predicts edges in an incremental way \emph{w.r.t} scale. We hence call the basic block in BDCN, which is constructed by inserting several SEMs into a VGG-type block, as the Incremental Detection Block (ID Block). This bi-directional cascade structure enforces each layer to focus on a specific scale, allowing for a more rational training procedure.

By combining SEM and BDCN, our method achieves consistent performance on three widely used datasets, \ie, \emph{BSDS500}, \emph{NYUDv2}, and \emph{Multicue}. It achieves ODS F-measure of 0.828, 1.3\% higher than current state-of-the art CED~\cite{Wang_2017_CVPR} on \emph{BSDS500}. It achieves 0.806 only using the trainval data of \emph{BSDS500} for training, and outperforms the human perception (ODS F-measure 0.803). To our best knowledge, we are the first that outperforms human perception by training only on trainval data of \emph{BSDS500}. Moreover, we achieve a better trade-off between model compactness and accuracy than existing methods relying on deeper models. With a shallow CNN structure, we obtain comparable performance with some well-known methods~\cite{bertasius2015high,shen2015deepcontour,Bertasius2015DeepEdge}. For example, we outperform HED~\cite{xie2015holistically} using only 1/6 of its parameters.
This shows the validity of our proposed SEM, which enriches the multi-scale representations in CNN. This work is also an original effort studying a rational training strategy for edge detection, \ie, employing the BDCN structure to train each CNN layer with layer-specific supervision.

\section{Related Work}\label{sec:related_work}
This work is related to edge detection, multi-scale representation learning, and network cascade structure. We briefly review these three lines of works, respectively.

\emph{{Edge Detection:}}
Most edge detection methods can be categorized into three groups, \ie, traditional edge operators, learning based methods, and the recent deep learning, respectively. Traditional edge operators~\cite{kittler1983accuracy,canny1986computational,torre1986edge,martin2004learning} detect edges by finding sudden changes in intensity, color, texture, \emph{etc}. Learning based methods spot edges by utilizing supervised models and hand-crafted features. For example, Doll\'ar \etal~\cite{dollar2015fast} propose  structured edge which jointly learns the clustering of groundtruth edges and the mapping of image patch to clustered token. Deep learning based methods use CNN to extract multi-level hierarchical features. Bertasius \etal~\cite{Bertasius2015DeepEdge} employ CNN to generate features of candidate contour points. Xie \etal~\cite{xie2015holistically} propose an end-to-end detection model that leverages the outputs from different intermediate layers with skip-connections. Liu \etal~\cite{Liu_2017_CVPR} further learn richer deep representations by concatenating features derived from all convolutional layers. Xu \etal~\cite{xu2017learning} introduce a hierarchical deep model to extract multi-scale features and a gated conditional random field to fuse them.

%Lim \etal~\cite{lim2013sketch} proposed to cluster contours into general tokens called Sketch Tokens, then train a random forest to map a local patch to these tokens.

\emph{{Multi-Scale Representation Learning:}}
Extraction and fusion of multi-scale features are fundamental and critical for many vision tasks, \emph{e.g.}, ~\cite{huang2017multi,yan2018hierarchical,chen2016attention}. Multi-scale representations can be constructed from multiple re-scaled images~\cite{farabet2013learning,pinheiro2014recurrent,eigen2015predicting}, \emph{i.e.}, an image pyramid, either by computing features independently at each scale~\cite{farabet2013learning} or using the output from one scale as the input to the next scale~\cite{pinheiro2014recurrent,eigen2015predicting}. Recently, innovative works DeepLab~\cite{chen2016deeplab} and PSPNet~\cite{zhao2017pyramid} use dilated convolutions and pooling to achieve multi-scale feature learning in image segmentation. Chen \etal~\cite{chen2016attention} propose an attention mechanism to softly weight the multi-scale features at each pixel location.

Like other image patterns, edges vary dramatically in scales. Ren \etal~\cite{ren2008multi} show that considering multi-scale cues does improve performance of edge detection. Multiple scale cues are also used in many approaches~\cite{witkin1987scale,ren2008multi,konishi2003statistical,xie2017ijcv,Liu_2017_CVPR,martin2004learning,xu2017learning}. Most of those approaches explore the scale-space of edges, \emph{e.g.}, using Gaussian smoothing at multiple scales~\cite{witkin1987scale} or extracting features from different scaled images~\cite{arbelaez2011contour}. Recent deep based methods employ image pyramid and hierarchal features. For example, Liu \etal~\cite{Liu_2017_CVPR} forward multiple re-scaled images to a CNN independently, then average the results. Our approaches follow a similar intuition, nevertheless, we build SEM to learn multi-scale representations in an efficient way, which avoids repetitive computation on multiple input images.

\emph{{Network Cascade:}}
Network cascade~\cite{ke2017srn,murthy2016deep,li2015convolutional,toshev2014deeppose,li2017not} is an effective scheme for many vision applications like classification~\cite{murthy2016deep}, detection~\cite{li2015convolutional}, pose estimation~\cite{toshev2014deeppose} and semantic segmentation~\cite{li2017not}. For example, Murthy \etal~\cite{murthy2016deep} treat easy and hard samples with different networks to improve classification accuracy. Yuan~\etal~\cite{yuan2016hard} ensemble a set of models with different complexities to process samples with different difficulties. Li \etal~\cite{li2017not} propose to classify easy regions in a shallow network and train deeper networks to deal with hard regions. Lin \etal~\cite{lin2017feature} propose a top-down architecture with lateral connections to propagate deep semantic features to shallow layers. Different from previous network cascade, BDCN is a bi-directional pseudo-cascade structure, which allows an innovative way to supervise each layer individually for layer-specific edge detection. To our best knowledge, this is an early and original attempt to adopt a cascade architecture in edge detection.

\section{Proposed Methods}

\subsection{Formulation}
\label{subsec: formulation}

Let ($X$, $Y$) denote one sample in the training set $\mathbb T$, where $X=\{x_j, j=1,\cdots,|X|\}$ is a raw input image and $Y=\{y_j, j=1,\cdots,|X|\}, y_j\in \{0,1\}$ is the corresponding groundtruth edge map. Considering the scale of edges may vary considerably in one image, we decompose edges in $Y$ into $S$ binary edge maps according to the scale of their depicted objects, \emph{i.e.},
\begin{equation} \label{eq:scl_decomp}
Y =\sum_{s=1}^{S} {Y_s},
\end{equation}
where ${Y_s}$ contains annotated edges corresponding to a scale $s$. Note that, we assume the scale of edges is in proportion to the size of their depicted objects.

Our goal is to learn an edge detector $\rm D(\cdot)$ capable of detecting edges at different scales. A natural way to design $\rm D(\cdot)$ is to train a deep neural network, where different layers correspond to different sizes of receptive field. Specifically, we can build a neural network $\rm N$ with $S$ convolutional layers. The pooling layers make adjacent convolutional layers depict image patterns at different scales.

For one training image $X$, suppose the feature map generated by the \emph{s}-th convolutional layer is $ {\rm N}_s(X) \in {\rm R}^{w\times h\times c}$. Using $ {\rm N}_s(X)$ as input, we design a detector ${\rm D}_s (\cdot)$ to spot edges at scale $s$. The training loss for ${\rm D}_s (\cdot)$ is formulated as
\begin{equation}\label{eq:loss_s}
{\mathcal L_s} = \sum\limits_{X \in \mathbb T} {|{P_s} - {Y_s}|},
\end{equation}
where ${P_s} = {\rm D}_s({{\rm N}_s}(X))$ is the edge prediction at scale $s$. The final detector $\rm D(\cdot)$ hence is derived as the ensemble of detectors learned from scale 1 to $S$.

To make the training with Eq.~\eqref{eq:loss_s} possible, $Y_s$ is required. It is not easy to decompose the groundtruth edge map $Y$ manually into different scales, making it hard to obtain the layer-specific supervision $Y_s$ for the $s$-th layer. A possible solution is to approximate $Y_s$ based on ground truth label $Y$ and edges predicted at other layers, \emph{i.e.},
\begin{equation} \label{eq:appro_one}
Y_s \sim Y - \sum\limits_{i \ne s} {P_i}.
\end{equation}
However, $Y_s$ computed in Eq.~\eqref{eq:appro_one} is not an
appropriate layer-specific supervision. In the following paragraph, we briefly explain the reason.

According to Eq.~\eqref{eq:appro_one}, for a training image, its predicted edges $P_s$ at layer $s$ should approximate $Y_s$, \emph{i.e.}, $P_s \sim Y - \sum \nolimits_{i
\ne s} {P_i}$. In other words, we can pass the other layers' predictions to layer $s$ for
training, resulting in an equivalent formulation, \emph{i.e.}, $Y \sim \sum \nolimits_{i}{P_i}$. The
training objective can thus become $\L = \L(\hat{Y}, Y)$, where $\hat{Y} = \sum_{i}P_i$.
The gradient \emph{w.r.t} the prediction $P_s$ of layer $s$ is
%$$%
\begin{equation}
\label{eq:eqderive}
\frac{\partial(L)}{\partial(P_s)} = \frac{\partial(L(\hat{Y}, Y))}{\partial(P_s)}
								  = \frac{\partial(L(\hat{Y},
Y))}{\partial(\hat{Y})}\cdot\frac{\partial(\hat{Y})}{\partial(P_s)}
								   .%\eqno{(4)}
%$$
\end{equation}%= \frac{\partial(\L(\hat{Y}, Y))}{\partial(\hat{Y})}
According to Eq.~\eqref{eq:eqderive}, for edge predictions $P_s$, $P_i$ at any two layers $s$ and $i$, $s \ne i$, their
loss gradients are equal because
$\frac{\partial(\hat{Y})}{\partial(P_s)}=\frac{\partial(\hat{Y})}{\partial(P_i)}=1$. This implies that,
with Eq.~\eqref{eq:appro_one}, the training process dose not necessarily differentiate the scales depicted by different layers, making it not appropriate for our layer-specific scale learning task.

To address the above issue, we approximate $Y_s$ with two complementary supervisions. One ignores the edges with scales smaller than $s$, and the other ignores the edges with larger scales. Those two supervisions train two edge detectors at each scale. We define those two supervisions at scale $s$ as
\begin{equation}
\label{eq:appro_two}
\begin{split}
Y_{s}^{s2d} = Y - \sum\limits_{i < s} {P_i}^{s2d}, \\
Y_{s}^{d2s} = Y - \sum\limits_{i > s} {P_i}^{d2s},
\end{split}
\end{equation}
where the superscript ${s2d}$ denotes information propagation from shallow layers to deeper layers, and ${d2s}$ denotes the prorogation from deep layers to shallower layers.

For scale $s$, the predicted edges ${P_s}^{s2d}$ and ${P_s}^{d2s}$ approximate to $Y_{s}^{s2d}$ and ${Y_s}^{d2s}$, respectively. Their combination is a reasonable approximation to $Y_{s}$, \emph{i.e.},
\begin{equation}\label{eq:bdcn}
{P_s}^{s2d} + {P_s}^{d2s} \sim 2Y-\sum\limits_{i < s} {P_i}^{s2d} - \sum\limits_{i > s} {P_i}^{d2s},
\end{equation}
where the edges predicted at scales $i \ne s$ are depressed. Therefore, we use ${P_s}^{s2d} + {P_s}^{d2s}$ to interpolate the edge prediction at scale $s$.

%We further modify existing deep neural networks to make it capable to train $\rm D_s(\cdot)$ with the loss function in Eq.~\eqref{eq:loss_s}. For example, we add the scale specific supervision $Y_s$ to the side output of $s$-th convolutional layer in $\rm N$.

Because different convolutional layers depict different scales, the depth of a neural network determines the range of scales it could model. A shallow network may not be capable to detect edges at all of the $S$ scales. However, a large number of convolutional layers involves too many parameters and makes the training difficult. To enable edge detection at different scales with a shallow network, we propose to enhance the multi-scale representation learned in each convolutional layer with the Scale Enhancement Module (SEM). The detail of SEM will be presented in Sec.~\ref{subsec: overall_arch}.

\subsection{Architecture of BDCN}
\label{subsec: overall_arch}
\begin{figure}[t]
\centerline{\epsfig{figure=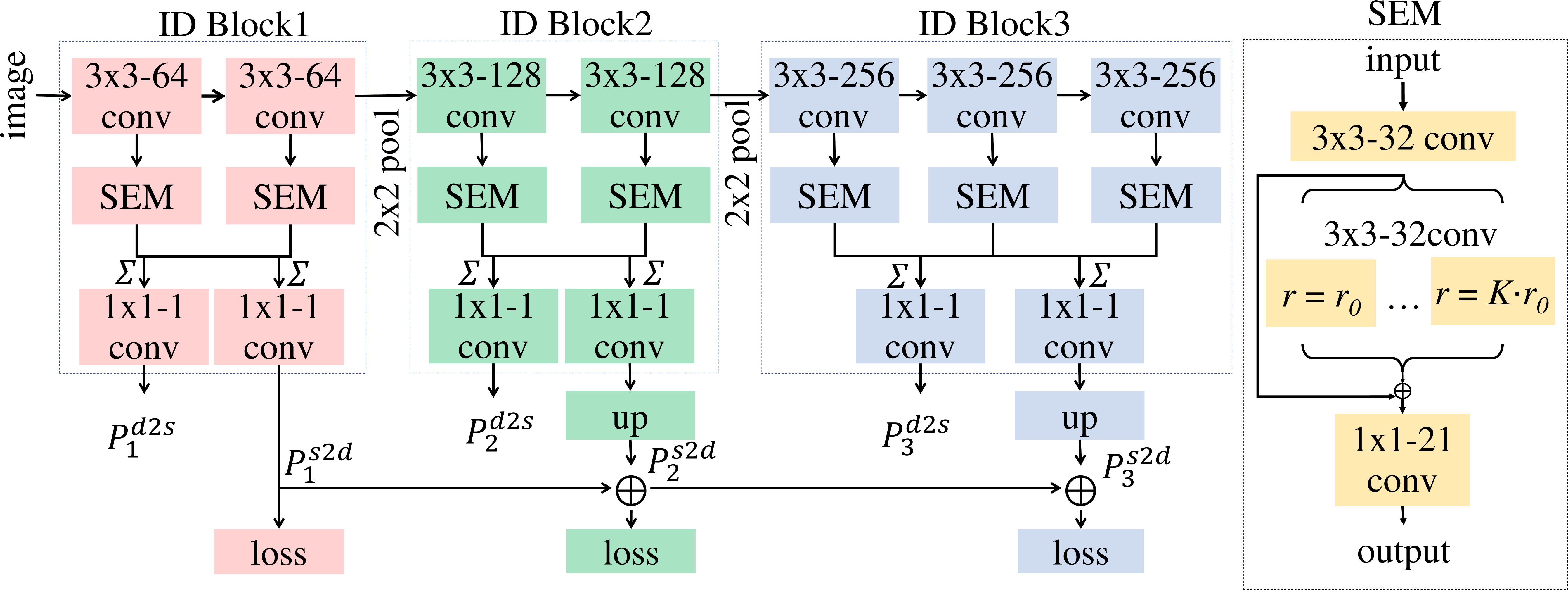, width=8.4cm}}
\caption{The detailed architecture of BDCN and SEM. For illustration, we only show 3 ID Blocks and the cascade from shallow to deep. The number of ID Blocks in our network can be flexibly set from 2 to 5 (see Fig.~\ref{fig:param-perf}).}
\label{fig:detail_bdcn}
\end{figure}

Based on Eq.~\eqref{eq:bdcn}, we propose a Bi-Directional Cascade Network (BDCN) architecture to achieve layer-specific training for edge detection. As shown in Fig.~\ref{fig:fig_overall}, our network is composed of multiple ID Blocks, each of which is learned with different supervisions inferred by a bi-directional cascade structure. Specifically, the network is based on the VGG16~\cite{simonyan2014very} by removing its three fully connected layers and last pooling layer. The 13 convolutional layers in VGG16 are then divided into 5 blocks, each follows a pooling layer to progressively enlarge the receptive fields in the next block. The VGG blocks evolve into ID Blocks by inserting several SEMs. We illustrate the detailed architecture of BDCN and SEM in Fig.~\ref{fig:detail_bdcn}.

{\textbf{ID Block}} is the basic component of our network. Each ID block produces two edge predictions. As shown in Fig.~\ref{fig:detail_bdcn}, an ID Block consists of several convolutional layers, each is followed by a SEM. The outputs of multiple SEMs are fused and fed into two 1$\times$1 convolutional layers to generate two edges predictions $P^{d2s}$ and $P^{s2d}$, respectively. The cascade structure shown in Fig.~\ref{fig:detail_bdcn} propagates the edge predictions from the shallow layers to deep layers. For the $s$-th block, $P_s^{s2d}$ is trained with supervision $Y_{s}^{s2d}$ computed in Eq.~\eqref{eq:appro_two}. $P_s^{d2s}$ is trained in a similar way. The final edge prediction is computed by fusing those intermediate edge predictions in a fusion layer using 1$\times$1 convolution.

\textbf{Scale Enhancement Module} is inserted into each ID Block to enrich the multi-scale representations in it. SEM is inspired by the dilated convolution proposed by Chen \etal~\cite{chen2016deeplab} for image segmentation. For an input two-dimensional feature map $\textbf{x}\in \mathcal{R}^{H\times W}$ with a convolution filter $\textbf{w}\in\mathcal{R}^{h\times w}$, the output $\textbf{y}\in\mathcal{R}^{H'\times W'}$ of dilated convolution at location $(i,j)$ is computed by
\begin{equation}
\textbf{y}_{ij}=\sum_{m,n}^{h,w} \textbf x_{[{i+r\cdot m},{j+r\cdot n}]}\cdot \textbf w_{[m,n]},
\label{eq:dilate}
\end{equation}
where $r$ is the dilation rate, indicating the stride for sampling input feature map. Standard convolution can be treated as a special case with $r=1$. Eq.~\eqref{eq:dilate} shows that dilated convolution enlarges the receptive field of neurons without reducing the resolution of feature maps or increasing the parameters.

As shown on the right side of Fig.~\ref{fig:detail_bdcn}, for each SEM we apply $K$ dilated convolutions with different dilation rates. For the $k$-th dilated convolution, we set its dilation rate as $r_k = max (1, r_0 \times k)$, which involves two parameters in SEM: the dilation rate factor $r_0$ and the number of convolution layers $K$. They are evaluated in Sec.~\ref{sec:parm_study}.

\subsection{Network Training}

Each ID Block in our network is trained with two layer-specific side supervisions. Besides that, we fuse the intermediate edge predictions with a fusion layer as the final result. Therefore, BDCN is trained with three types of loss. We formulate the overall loss $\mathcal{L}$ as,
\begin{alignat}{2}
\label{eq:object}
\mathcal{L} &= w_{side}\cdot\mathcal{L}_{side} + w_{fuse}\cdot\mathcal{L}_{fuse}(P, Y),\\
\mathcal{L}_{side} &= \sum_{s=1}^{S}\mathcal{L}(P_{s}^{d2s}, Y_s^{d2s}) + \mathcal{L}(P_{s}^{s2d}, Y_s^{s2d}),
\end{alignat}
where $w_{side}$ and $w_{fuse}$ are weights for the side loss and fusion loss, respectively. $P$ denotes the final edge prediction.

The function $\mathcal{L}(\cdot)$ is computed at each pixel with respect to its edge annotation. Because the distribution of edge/non-edge pixels is heavily biased, we employ a class-balanced cross-entropy loss as $\mathcal{L}(\cdot)$. Because of the inconsistency of annotations among different annotators, we also introduce a threshold $\gamma$ for loss computation. For a groudtruth $Y=(y_j, j=1,...,|Y|), y_j\in(0,1)$, we define $Y_+=\{y_j,y_j>\gamma\}$ and $Y_-=\{y_j,y_j=0\}$. Only pixels corresponding to $Y_+$ and $Y_-$ are considered in loss computation. We hence define $\mathcal{L}(\cdot)$ as
\begin{equation}
\mathcal{L}\left(\hat{Y}, Y\right) = -\alpha\sum_{j\in Y_-}log(1 - \hat{y}_j)-\beta\sum_{j\in Y_+}log(\hat{y}_j),
\end{equation}
where $\hat{Y}=(\hat{y}_j, j=1,...,|\hat{Y}|), \hat{y}_j\in(0,1)$ denotes a predicted edge map, $\alpha = \lambda\cdot|Y_+|/(|Y_+|+|Y_-|), \beta = |Y_-|/(|Y_+|+|Y_-|)$ balance the edge/non-edge pixels. $\lambda$ controls the weight of positive over negative samples.

\begin{figure}[t]
\centerline{\epsfig{figure=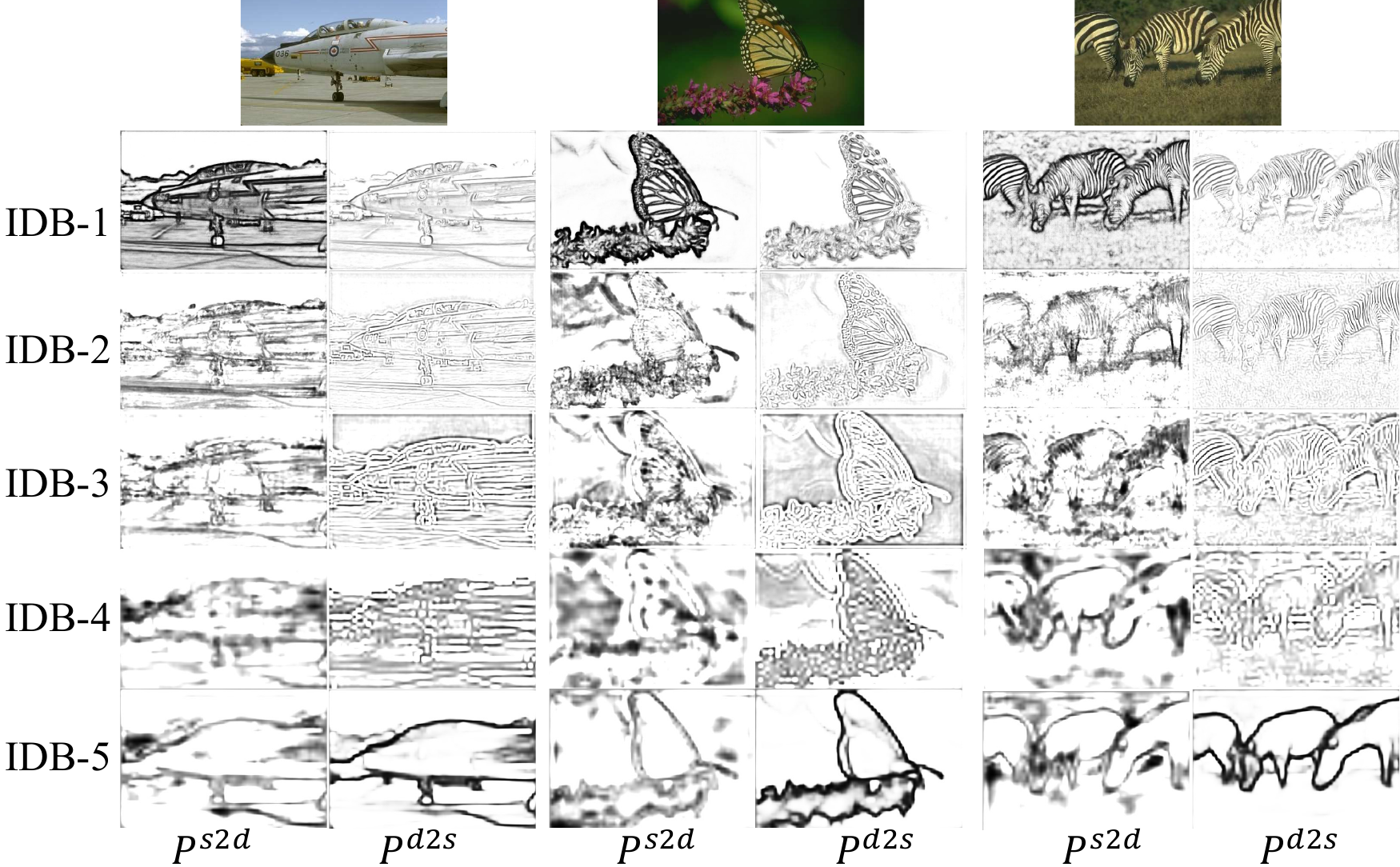, width=8.4cm}}
\caption{Examples of edges detected by different ID Blocks (IDB for short). Each ID Block generates two edge predictions, $P^{s2d}$ and $P^{d2s}$, respectively.}
\label{fig:scale_dif}
\end{figure}

Fig.~\ref{fig:scale_dif} shows edges detected by different ID blocks. We observe that, edges detected by different ID Blocks correspond to different scales. The shallow ID Blocks produce strong responses on local details and deeper ID Blocks are more sensitive to edges at larger scale. For instance, detailed edges on the body of zebra and butterfly can be detected by shallow ID Block, but are depressed by deeper ID Block. The following section tests the validity of BDCN and SEM.

\section{Experiments}
\label{sec:sec_exp}
\subsection{Datasets}

We evaluate the proposed approach on three public datasets: \emph{BSDS500}~\cite{arbelaez2011contour}, \emph{NYUDv2}~\cite{silberman2012indoor}, and \emph{Multicue}~\cite{mely2016systematic}.

\emph{{BSDS500}} contains 200 images for training, 100 images for validation, and 200 images for testing. Each image is manually annotated by multiple annotators. The final groundtruth is the averaged annotations by the annotators. We also utilize the strategies in~\cite{xie2015holistically,Liu_2017_CVPR,Wang_2017_CVPR} to augment training and validation sets by randomly flipping, scaling and rotating images. Following those works, we also adopt the PASCAL VOC Context dataset~\cite{Mottaghi2014cvpr} as our training set.

\emph{{NYUDv2}} consists of 1449 pairs of aligned RGB and depth images. It is split into 381 training, 414 validation, and 654 testing images. \emph{NYUDv2} is initially used for scene understanding, hence is also used for edge detection in previous works~\cite{gupta2013perceptual,xiaofeng2012discriminatively,xie2015holistically,Liu_2017_CVPR}. Following those works, we augment the training set by randomly flipping, scaling, and rotating training images.

\emph{{Multicue}}~\cite{mely2016systematic} contains 100 challenging natural scenes. Each scene has two frame sequences taken from left and right view, respectively. The last frame of left-view sequence is annotated with edges and boundaries. Following~\cite{mely2016systematic,Liu_2017_CVPR,xie2017ijcv}, we randomly split 100 annotated frames into 80 and 20 images for training and testing, respectively. We also augment the training data with the same way in~\cite{xie2015holistically}.

\subsection{Implementation Details}

We implement our network using PyTorch. The VGG16~\cite{simonyan2014very} pretrained on ImageNet~\cite{deng2009imagenet} is used to initialize the backbone. The threshold $\gamma$ used for loss computation is set as 0.3 for \emph{BSDS500}. $\gamma$ is set as 0.3 and 0.4 for \emph{Multicue} \emph{boundary} and \emph{edges} datasets, respectively. \emph{NYUDv2} provides binary annotations, thus does not need to set $\gamma$ for loss computation. Following~\cite{Liu_2017_CVPR}, we set the parameter $\lambda$ as 1.1 for \emph{BSDS500} and \emph{Multicue}, set $\lambda$ as 1.2 for \emph{NYUDv2}.

SGD optimizer is adopted to train our network. On \emph{BSDS500} and \emph{NYUDv2}, we set the batch size to 10 for all the experiments. The initial learning rate, momentum and weight decay are set to 1$e$-6, 0.9, and 2e-4 respectively. The learning rate decreases by 10 times after every 10k iterations. We train 40k iterations for \emph{BSDS500} and \emph{NYUDv2}, 2k and 4k iterations for \emph{Multicue} \emph{boundary} and \emph{edge}, respectively. $w_{side}$ and $w_{fuse}$ are set as 0.5, and 1.1, respectively. Since \emph{Multicue} dataset includes high resolution images, we randomly crop 500$\times$500 patches from each image in training. All the experiments are conducted on a NVIDIA GeForce1080Ti GPU with 11GB memory.

We follow previous works~\cite{xie2015holistically,Liu_2017_CVPR,Wang_2017_CVPR,xu2017learning}, and perform standard Non-Maximum Suppression (NMS) to produce the final edge maps. For a fair comparison with other work, we report our edge detection performance with commonly used evaluation metrics, including Average Precision (AP), as well as F-measure at both Optimal Dataset Scale (ODS) and Optimal Image Scale (OIS). The maximum tolerance allowed for correct matches between edge predictions and groundtruth annotations is set to 0.0075 for \emph{BSDS500} and \emph{Multicue} dataset, and is set to 0.011 for \emph{NYUDv2} dataset as in previous works~\cite{Liu_2017_CVPR,mely2016systematic,xie2017ijcv}.

\subsection{Ablation Study}\label{sec:parm_study}

\begin{table}[t]
\begin{center}
\label{tab:parmstudy}
\caption{Impact of SEM parameters to the edge detection performance on \emph{BSDS500} validation set. (a) shows the impact of $K$ with $r_0$=4. (b) shows the impact of $r_0$ with $K$=3.}
\vspace{4mm}
\begin{minipage}{4cm}
\label{tab:ablation_k}
\setlength{\tabcolsep}{3pt}
\centering (a)
\vspace{-1mm}
\begin{center}
\scriptsize
\begin{tabular}{c||c|c|c}
\hline
$K$ & ODS & OIS & AP \\
\hline\hline
0 & .7728 & .7881 & .8093\\
1 & .7733 & .7845 & .8139\\
2 & .7738 & .7876 & .8169\\
3 & \textbf{.7748} & \textbf{.7894} & \textbf{.8170}\\
4 & .7745 & .7896 & .8166\\
\hline
\end{tabular}
\end{center}
\end{minipage}
%\hfill
\hspace{0.0cm}
\begin{minipage}{4cm}
\label{tab:ablation_rate}
\setlength{\tabcolsep}{3pt}
\centering (b)
\vspace{-1mm}
%\vspace{-0.3cm}
\begin{center}
\scriptsize
\begin{tabular}{c|c||c|c|c}
\hline
$r_0$ & rate & ODS & OIS & AP \\
\hline
\hline
0 & 1,1,1 & .7720 & .7881 & .8116 \\
1 & 1,2,3 & .7721 & .7882 & .8124 \\
2 & 2,4,6 & .7725 & .7875 & .8132 \\
4 & 4,8,12 & \textbf{.7748} & \textbf{.7894} & \textbf{.8170}  \\
8 & 8,16,24 & .7742 & .7889 & .8169 \\
\hline
\end{tabular}
\end{center}
\end{minipage}
\end{center}
\end{table}

\begin{table}[h]
\caption{Validity of components in BDCN on \emph{BSDS500} validation set. (a)  tests different cascade architectures. (b) shows the validity of SEM and the bi-directional cascade architecture.}
\vspace{4mm}
\begin{minipage}{4cm}
%\caption{Quantitative results of different cascade policies on BSDS500.}
\centering (a)
\vspace{-3mm}
\begin{center}
\label{tab:ablation_cascade}
\setlength{\tabcolsep}{2.5pt}
\scriptsize
\begin{tabular}{c||c c c}
\hline
Architecture & ODS & OIS & AP \\
\hline\hline
baseline & .7681 & .7751 & .7912 \\
\hline
S2D & .7683 & .7802 & .7978 \\
D2S & .7710 & .7816 & \textbf{.8049}\\
\hline
%S2D+D2S & \textbf{.7762} & \textbf{.7872} & .8013\\
S2D+D2S & \multirow{2}{*}{\textbf{.7762}}  & \multirow{2}{*}{\textbf{.7872}} & \multirow{2}{*}{.8013} \\
(BDCN w/o SEM)&\\
\hline
\end{tabular}
\end{center}
\end{minipage}
\hfill
\begin{minipage}{4cm}
\label{tab:ablation_c}
\centering (b)
\vspace{-3mm}
\begin{center}
\setlength{\tabcolsep}{2.5pt}
\scriptsize
\begin{tabular}{c||c c c}
\hline
Method & ODS & OIS & AP \\
\hline \hline
baseline & .7681 & .7751 & .7912 \\
\hline
SEM & .7748 & \textbf{.7894} & \textbf{.8170} \\
S2D+D2S & \multirow{2}{*}{.7762}  & \multirow{2}{*}{.7872} & \multirow{2}{*}{.8013} \\
(BDCN w/o SEM)&\\
\hline
BDCN & \textbf{.7765} & .7882 &.8091 \\
\hline
\end{tabular}
\end{center}
\end{minipage}
%\vindent
\end{table}

In this section,
we conduct experiments on \emph{BSDS500} to study the impact of parameters and verify each component in our network. We train the network on the \emph{BSDS500} training set and evaluate on the validation set. Firstly, we test the impact of the parameters in SEM, \emph{i.e.}, the number of dilated convolutions $K$ and the dilation rate factor $r_0$. Experimental results are summarized in Table \ref{tab:ablation_k}.

Table \ref{tab:ablation_k} (a) shows the impact of $K$ with $r_0$=4. Note that, $K$=0 means directly copying the input as output. The results demonstrate that setting $K$ larger than 1 substantially improves the performance. However, too large $K$ does not constantly boost the performance. The reason might be that, large $K$ produces high dimensional outputs and makes edge extraction from such high dimensional data difficult. Table \ref{tab:ablation_k} (b) also shows that larger $r_0$ improves the performance. But the performance starts to drop with too large $r_0$, \emph{e.g.}, $r_0$=8. In our following experiments, we fix $K$=3 and $r_0$=4.

Table \ref{tab:ablation_cascade} (a) shows the comparison among different cascade architectures, \emph{i.e.}, single direction cascade from shallow to deep layers (S2D), from deep to shallow layers (D2S), and the bi-directional cascade (S2D+D2S), \emph{i.e.}, the BDCN w/o SEM. Note that, we use the VGG16 network without fully connected layer as baseline. It can be observed that, both S2D and D2S structures outperform the baseline. This shows the validity of the cascade structure in network training. The combination of these two cascade structures,\emph{ i.e.}, S2D+D2S, results in the best performance. We further test the performance of combining SEM and S2D+D2S and summarize the results in Table~\ref{tab:ablation_c} (b), which shows that SEM and bi-directional cascade structure consistently improve the performance of baseline, \emph{e.g.}, improve the ODS F-measure by 0.7\% and 0.8\% respectively. Combining SEM and S2D+D2S results in the best performance. We can conclude that, the components introduced in our method are valid in boosting edge detection performance.

\subsection{Comparison with Other Works}
\label{subsec:res}

\begin{table}
\begin{center}
%\captionsetup{type=table}
\caption{Comparison with other methods  on \emph{BSDS500} test set. \dag indicates trained with additional PASCAL-Context data. \ddag indicates the fused result of multi-scale images.}
\vspace{1mm}
\label{tab:bsds500_tab}
\scriptsize
\setlength{\tabcolsep}{5.4pt}
\begin{tabular}{c||c c c c}
  \hline
  Method & ODS & OIS & AP \\
  \hline
  \hline
  Human & .803 & .803 & -- \\
  \hline
  %MShift \cite{comaniciu2002mean} & .598 & .645 & -- \\
  %Canny \cite{canny1986computational} & .611 & .676 & -- \\
  %NCut \cite{shi2000normalized}
  %EGB \cite{felzenszwalb2004efficient} & .614 & .658 & -- \\
  %\hline
  %Mean Shift \cite{comaniciu2002mean} & .640 & .680 & .560  \\
  %\hline
  %ISCRA \cite{ren2013image} & .724 & .752 & .783 \\
  %gPb-UCM \cite{arbelaez2011contour} & .729 & .755 & .696 \\
  %\hline
  SCG \cite{xiaofeng2012discriminatively} & .739 & .758 & .773 \\
  PMI \cite{isola2014crisp} & .741 & .769 & .799 \\
  %MCG \cite{arbelaez2014multiscale} & .744 & .
  OEF \cite{hallman2015oriented} & .746 & .770 & .820 \\
  \hline
  DeepContour \cite{shen2015deepcontour} & .757 & .776 & .800 \\
  HFL \cite{bertasius2015high} & .767 & .788 & .795 \\
  HED \cite{xie2015holistically} & .788 & .808 & .840 \\
  CEDN \cite{yang2016object} \dag & .788 & .804 & -- \\
  COB \cite{cob16a} & .793 & .820 & .859 \\
  DCD \cite{liao2017deep} & .799 & .817 & .849 \\
  AMH-Net \cite{xu2017learning} & .798 & .829 & .869 \\
  \hline
  RCF \cite{Liu_2017_CVPR} & .798 & .815 & -- \\
  RCF \cite{Liu_2017_CVPR} \dag & .806 & .823 & -- \\
  RCF \cite{Liu_2017_CVPR} \ddag & .811 & .830 & -- \\
  \hline
  Deep Boundary \cite{kokkinos2015pushing} & .789 & .811 & .789 \\
  Deep Boundary \cite{kokkinos2015pushing} \ddag & .809 & .827 & .861 \\
  Deep Boundary \cite{kokkinos2015pushing} \ddag ~+~ Grouping& .813 & .831 & .866 \\
  \hline
  CED \cite{Wang_2017_CVPR} & .794 & .811 & .847 \\
  CED \cite{Wang_2017_CVPR} \ddag & .815 & .833 & .889 \\
  \hline
  LPCB \cite{deng2018learning} & .800 & .816 & -- \\
  LPCB \cite{deng2018learning} \dag & .808 & .824 & -- \\
  LPCB \cite{deng2018learning} \ddag & .815 & .834 & -- \\
  \hline
  BDCN & \textbf{.806} & \textbf{.826} & .847 \\
  BDCN \dag & \textbf{.820} & \textbf{.838 }& \textbf{.888} \\
  BDCN \ddag & \textbf{.828} & \textbf{.844} & \textbf{.890} \\
\hline
\end{tabular}
\end{center}
\end{table}

\begin{figure}[t]
\begin{center}
\centering
\centerline{\epsfig{figure=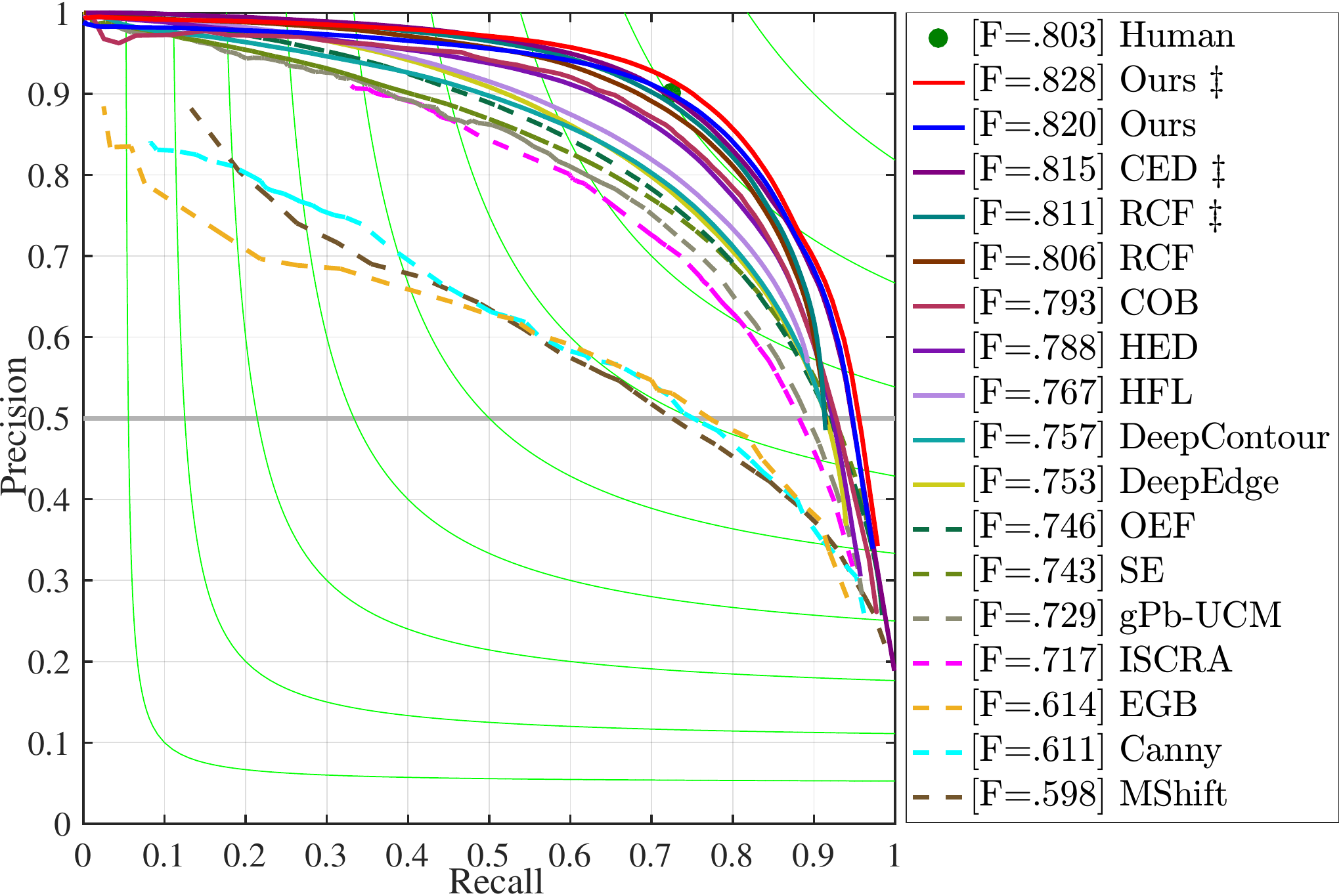,width=8.3cm}}
\caption{The precision-recall curves of our method and other works on \emph{BSDS500} test set.}
\label{fig:bsds_pr}
\vspace{-0.3cm}
\end{center}
\end{figure}

\emph{Performance on BSDS500:}
%note................ whether to compare traditional methods
We compare our approach with recent deep learning based methods including CED~\cite{Wang_2017_CVPR}, RCF~\cite{Liu_2017_CVPR}, DeepBoundary~\cite{kokkinos2015pushing}, DCD~\cite{liao2017deep}, COB~\cite{cob16a}, HED~\cite{xie2015holistically}, HFL~\cite{bertasius2015high}, DeepEdge~\cite{Bertasius2015DeepEdge} and DeepContour~\cite{shen2015deepcontour}, and traditional edge detection methods, including SCG~\cite{xiaofeng2012discriminatively}, PMI~\cite{isola2014crisp} and OEF~\cite{hallman2015oriented}. The comparison on \emph{BSDS500} is summarized in Table \ref{tab:bsds500_tab} and Fig. \ref{fig:bsds_pr}, respectively.

As shown in the results, our method obtains the F-measure ODS of 0.820 using single scale input, and achieves 0.828 with multi-scale inputs, both outperform all of these competing methods. Using a single-scale input, our method still outperforms the recent CED~\cite{Wang_2017_CVPR} and DeepBoundary~\cite{kokkinos2015pushing} that use multi-scale inputs. Our method also outperforms the human perception by 2.5\% in F-measure ODS. The F-measure OIS and AP of our approach are also higher than the ones of the other methods.

%\begin{table}
%\begin{center}
%\scriptsize
%\caption{Performance that only trained on the trainval data of BSDS500.}
%\begin{tabular}{c||c c c}
%\hline
%Methods & ODS & OIS & AP \\
%\hline
%\hline
%RCF~\cite{Liu_2017_CVPR} & 0.798 & 0.815 & -- \\
%CED~\cite{Wang_2017_CVPR} & 0.794 & 0.811 & 0.847 \\
%DeepBoundary~\cite{kokkinos2015pushing} & 0.789 & 0.829 & 0.869 \\
%AMH-Net~\cite{xu2017learning} & 0.798 & 0.829 & 0.869 \\
%\hline
%BED & \textbf{0.806} & \textbf{0.826} & \textbf{0.847} \\
%BED\dag & \textbf{0.815} & \textbf{0.832} & \textbf{0.871} \\
%\hline
%\end{tabular}
%\end{center}
%\end{table}

\begin{table}[t]
\begin{center}
%\captionsetup{type=table}
\caption{Comparison with recent works  on \emph{NYUDv2}.}
\vspace{1mm}
\scriptsize
\label{tab:nyud_tab}
\setlength{\tabcolsep}{3pt}
\begin{tabular}{c|c||c c c}
  \hline
  \multicolumn{2}{c||}{Method}  & ODS & OIS & AP \\
  \hline\hline
  gPb-UCM \cite{arbelaez2011contour} & \multirow{5}{*}{RGB} & .632 & .661 & .562 \\
  gPb+NG \cite{gupta2013perceptual} & &.687 & .716 & .629 \\
  OEF\cite{hallman2015oriented} & &.651 & .667 & -- \\
  SE \cite{dollar2015fast} & &.695 & .708 & .679 \\
  SE+NG+ \cite{guptaECCV14} & &.706 & .734 & .738 \\
  \hline
  \multirow{3}{*}{HED \cite{xie2015holistically}}& RGB & .720 & .734 & .734 \\
  & HHA  & .682 & .695 & .702 \\
  & RGB-HHA  & .746 & .761 & .786 \\
  \hline
  \multirow{3}{*}{RCF \cite{Liu_2017_CVPR}} & RGB & .729 & .742 & -- \\
  & HHA & .705 & .715 & -- \\
  & RGB-HHA & .757 & .771 & -- \\
  \hline
  \multirow{3}{*}{AMH-Net-ResNet50 \cite{xu2017learning}}& RGB & .744 & .758 & .765 \\
  & HHA & .716 & .729 & .734 \\
  & RGB-HHA & .771 & .786 & .802 \\
  \hline
  \multirow{3}{*}{LPCB \cite{deng2018learning}} & RGB & .739 & .754 & -- \\
  & HHA & .707 & .719 &  -- \\
  & RGB-HHA & .762 & .778 & -- \\
  \hline
  COB-ResNet50\cite{maninis2018convolutional} & RGB-HHA & .784 & .805 & 825 \\
  \hline
  \multirow{3}{*}{BDCN} & RGB & \textbf{.748} & \textbf{.763} &\textbf{ .770} \\
  & HHA & .707 & .719 & .731 \\
  & RGB-HHA & .765 & .781 & .813\\
  \hline
\end{tabular}
\end{center}
\end{table}

\begin{figure}[t]
\begin{center}
\centerline{\epsfig{figure=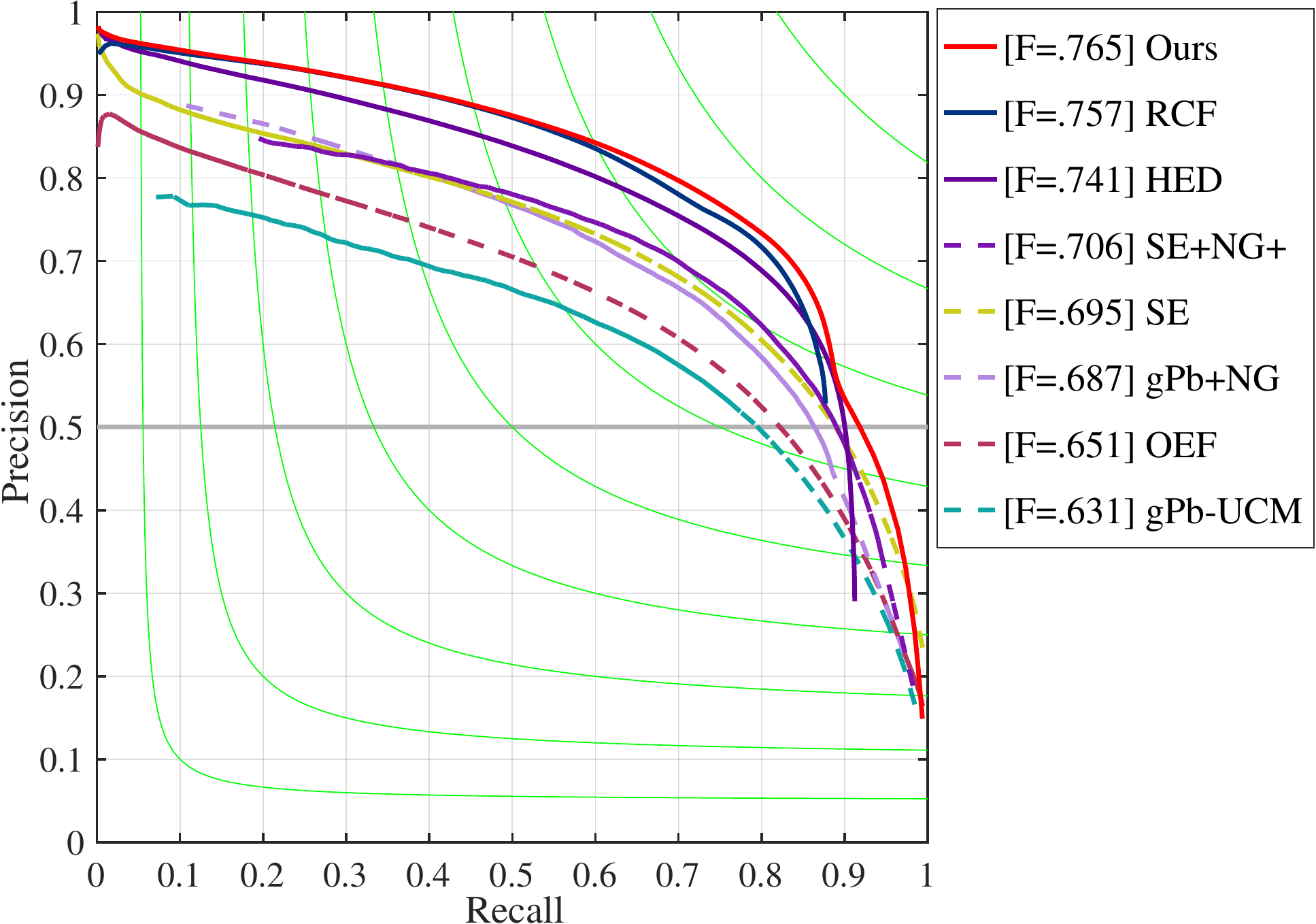,width=8.2cm}}
\caption{The precision-recall curves of our method and compared works on \emph{NYUDv2}.}
\label{fig:nyud_pr}
\end{center}
\end{figure}

\emph{Performance on NYUDv2:}
\emph{NYUDv2} has three types of inputs, \ie, RGB, HHA, and RGB-HHA, respectively. Following previous works~\cite{xie2015holistically,Liu_2017_CVPR}, we perform experiments on all of them. The results of RGB-HHA are obtained by averaging the edges detected on RGB and HHA. Table \ref{tab:nyud_tab} shows the comparison of our method with several recent approaches, including gPb-ucm~\cite{arbelaez2011contour}, OEF~\cite{hallman2015oriented}, gPb+NG~\cite{gupta2013perceptual}, SE+NG+~\cite{guptaECCV14}, SE~\cite{dollar2015fast}, HED~\cite{xie2015holistically}, RCF~\cite{Liu_2017_CVPR} and AMH-Net~\cite{xu2017learning}. Fig.~\ref{fig:nyud_pr} shows the precision-recall curves of our method and other competitors. All of the evaluation results are based on a single scale input.

As shown in Table~\ref{tab:nyud_tab} and Fig.~\ref{fig:nyud_pr}, our performance is competitive, \emph{i.e.}, outperforms most of the compared works except AMH-Net~\cite{xu2017learning}. Note that, AMH-Net applies the deeper ResNet50 to construct the edge detector. With a shallower network, our method still outperforms AMH-Net on the RGB image, \emph{i.e.}, our 0.748 vs. 0.744 of AMH-Net in F-measure ODS. Compared with previous works, our improvement over existing works is actually more substantial, \emph{e.g.}, on \emph{NYUDv2} our gains over RCF~\cite{Liu_2017_CVPR} and HED~\cite{xie2015holistically} are 0.019 and 0.028 in ODS, higher than the 0.009 gain of RCF~\cite{Liu_2017_CVPR} over HED~\cite{xie2015holistically}.

\begin{table}[t]
\begin{center}
\caption{Comparison with recent works on \emph{Multicue}. \ddag indicates the fused result of multi-scale images.} \label{tab:multicue}
\vspace{1mm}
\scriptsize
\setlength{\tabcolsep}{3pt}
\begin{tabular}{c|c|c|c|c}
\Xhline{0.8pt}
  Cat. & Method & ODS & OIS & AP \\
%\hline
%\hline
\Xhline{0.8pt}
\multirow{6}{*}{Boundary} & Human \cite{mely2016systematic} & .760 (0.017) & -- & -- \\
 & Multicue \cite{mely2016systematic} & .720 (0.014) & -- & --  \\
 & HED \cite{xie2017ijcv} & .814 (0.011) & .822 (0.008) & .869(0.015) \\
 & RCF \cite{Liu_2017_CVPR} & .817 (0.004) & .825 (0.005) & -- \\
 & RCF \cite{Liu_2017_CVPR} \ddag& .825 (0.008) & .836 (0.007) & -- \\
 & BDCN & .836 (\textbf{0.001}) & .846(\textbf{0.003}) & .893(\textbf{0.001}) \\
 & BDCN \ddag & \textbf{.838}(0.004) & \textbf{.853}(0.009) & \textbf{.906}(0.005) \\
\hline
\multirow{6}{*}{Edge} & Human \cite{mely2016systematic} & .750 (0.024) & -- & -- \\
 & Multicue \cite{mely2016systematic} & .830 (0.002) & -- & --\\
 & HED \cite{xie2017ijcv} & .851 (0.014) & .864 (0.011)  & --\\
 & RCF \cite{Liu_2017_CVPR} & .857 (0.004) & .862 (0.004)  & --\\
 & RCF \cite{Liu_2017_CVPR} \ddag& .860 (0.005) & .864 (0.004) & -- \\
 & BDCN & .891 (\textbf{0.001}) & .898 (\textbf{0.002}) & {.935}(\textbf{0.002}) \\
 & BDCN \ddag & \textbf{.894}(0.002) & \textbf{.901}(0.004) & \textbf{.941}(0.005) \\
\hline
\end{tabular}
\end{center}
\end{table}

\begin{figure}[t]
\centering
\centerline{\epsfig{figure=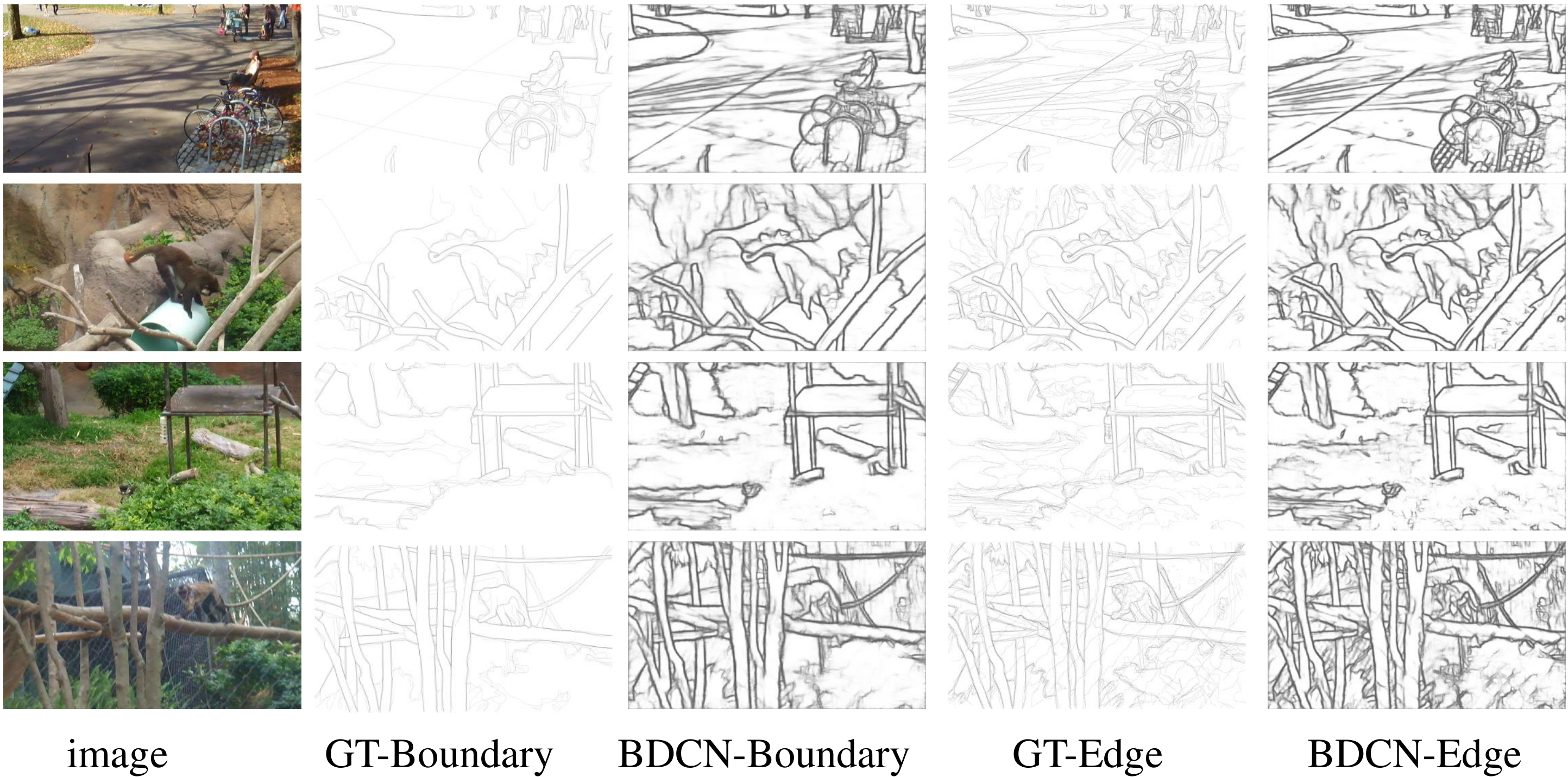,width=8.3cm}}
\caption{Examples of our edge detection results before Non-Maximum Suppression on \emph{Multicue} dataset.}
\label{fig:fig_multi_cue}
\end{figure}

\emph{Performance on Multicue:}
\emph{Multicue} consists of two sub datasets, \ie, \emph{Multicue boundary} and \emph{Multicue edge}. As done in RCF~\cite{Liu_2017_CVPR} and the recent version of HED~\cite{xie2017ijcv}, we average the scores of three independent experiments as the final result. We show the comparison with recent works in Table \ref{tab:multicue}, where our method achieves substantially higher performance than RCF~\cite{Liu_2017_CVPR} and HED~\cite{xie2015holistically}. For boundary detection task, we outperform RCF and HED by 1.3\% and 2.4\%, respectively in F-measure ODS. For edge detection task, our performance is 3.4\% and 4.3\% higher than the ones of RCF and HED. Moreover, the performance fluctuation of our method is considerably smaller than those two methods, which means our method delivers more stable results. Some edge detection results generated by our approach on \emph{Multicue} are presented in Fig. \ref{fig:fig_multi_cue}.

\begin{figure}[t]
\centering
\centerline{\epsfig{figure=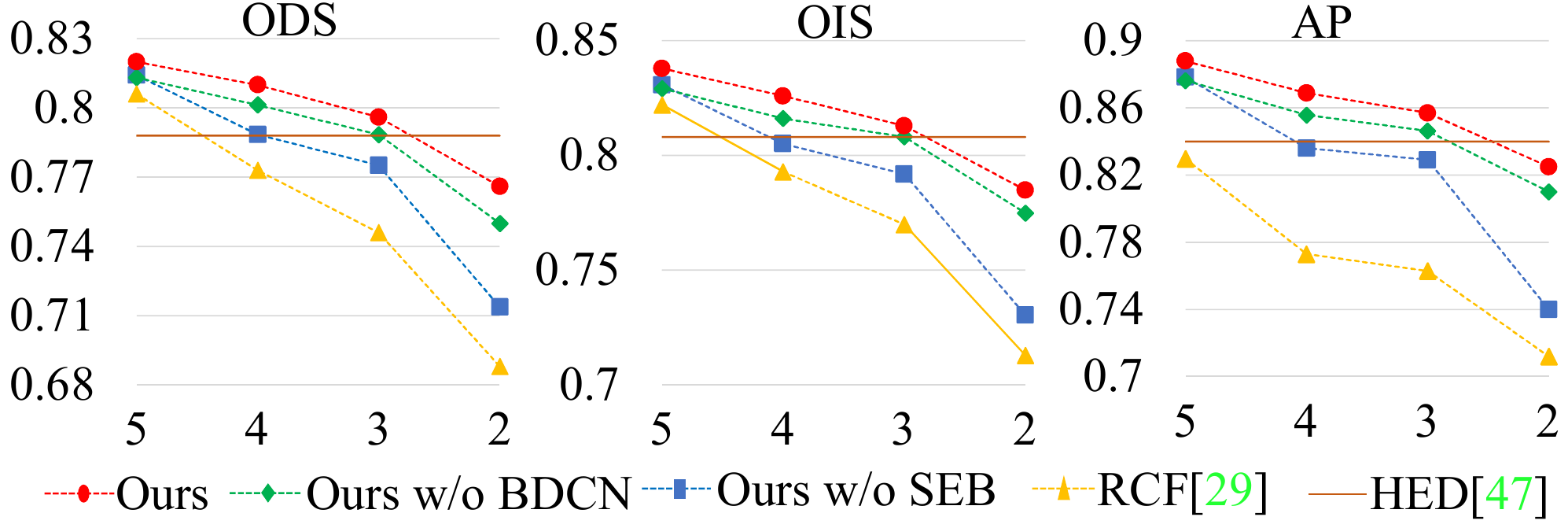, width=8.5cm}}
\caption{Comparison of edge detection accuracy as we decrease the number of ID Blocks from 5 to 2. HED learned with VGG16 is denoted as the solid line for comparison. }
\label{fig:ablation}
\end{figure}

\begin{figure}
\begin{center}
\centerline{\epsfig{figure=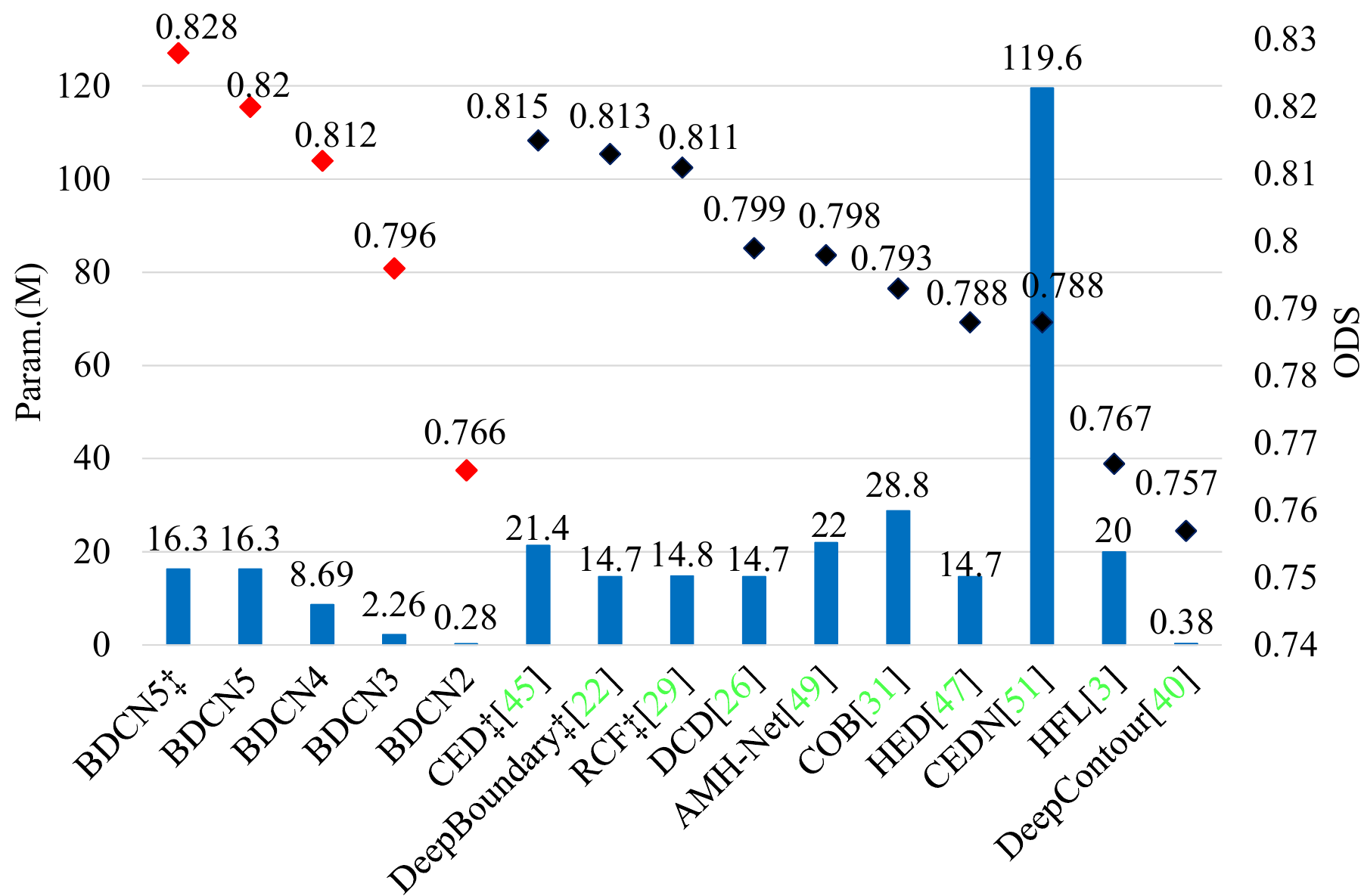,width=8.5cm}}
\caption{Comparison of parameters and performance with other methods. The number behind ``BDCN'' indicates the number of ID Block. \ddag means the multiscale results. }
\label{fig:param-perf}
\vspace{-2mm}
\end{center}
\end{figure}

\emph{Discussions:} The above experiments have shown the competitive performance of our proposed method. We further test the capability of our method in learning multi-scale representations with shallow networks. We test our approach and RCF with different depth of networks, \ie, using different numbers of convolutional block to construct the edge detection model. Fig.~\ref{fig:ablation} presents the results on \emph{BSDS500}. As shown in Fig. \ref{fig:ablation}, the performance of RCF~\cite{Liu_2017_CVPR} drops more substantially than our method as we decrease the depth of networks. This verifies that our approach is more effective in detecting edges with shallow networks. We also show the performance of our approach without the SEM and the BDCN structure. These ablations show that removing either BDCN or SEM degrades the performance. It is also interesting to observe that, without SEM, the performance of our method drops substantially. This hence verifies the importance of SEM to multi-scale representation learning in shallow networks.

Fig.~\ref{fig:param-perf} further shows the comparison of parameters \emph{vs}. performance of our method with other deep net based methods on \emph{BSDS500}. With 5 convolutional blocks in VGG16, HED~\cite{xie2015holistically}, RCF~\cite{Liu_2017_CVPR}, and our method use similar number of parameters, \emph{i.e.}, about 16M. As we decrease the number of ID Blocks from 5 to 2, our number of parameters decreases dramatically, drops to 8.69M, 2.26M, and 0.28M, respectively. Our method still achieves F-measure ODS of 0.766 using only two ID Blocks with 0.28M parameters. It also outperforms HED and RCF with a more shallow network, \emph{i.e.}, with 3 and 4 ID Blocks respectively. For example, it outperforms HED by 0.8\% with 3 ID Blocks and just 1/6 parameters of HED. We thus conclude that, our method can achieve promising edge detection accuracy even with a compact shallow network.

\begin{figure}[t]
\centering
\centerline{\epsfig{figure=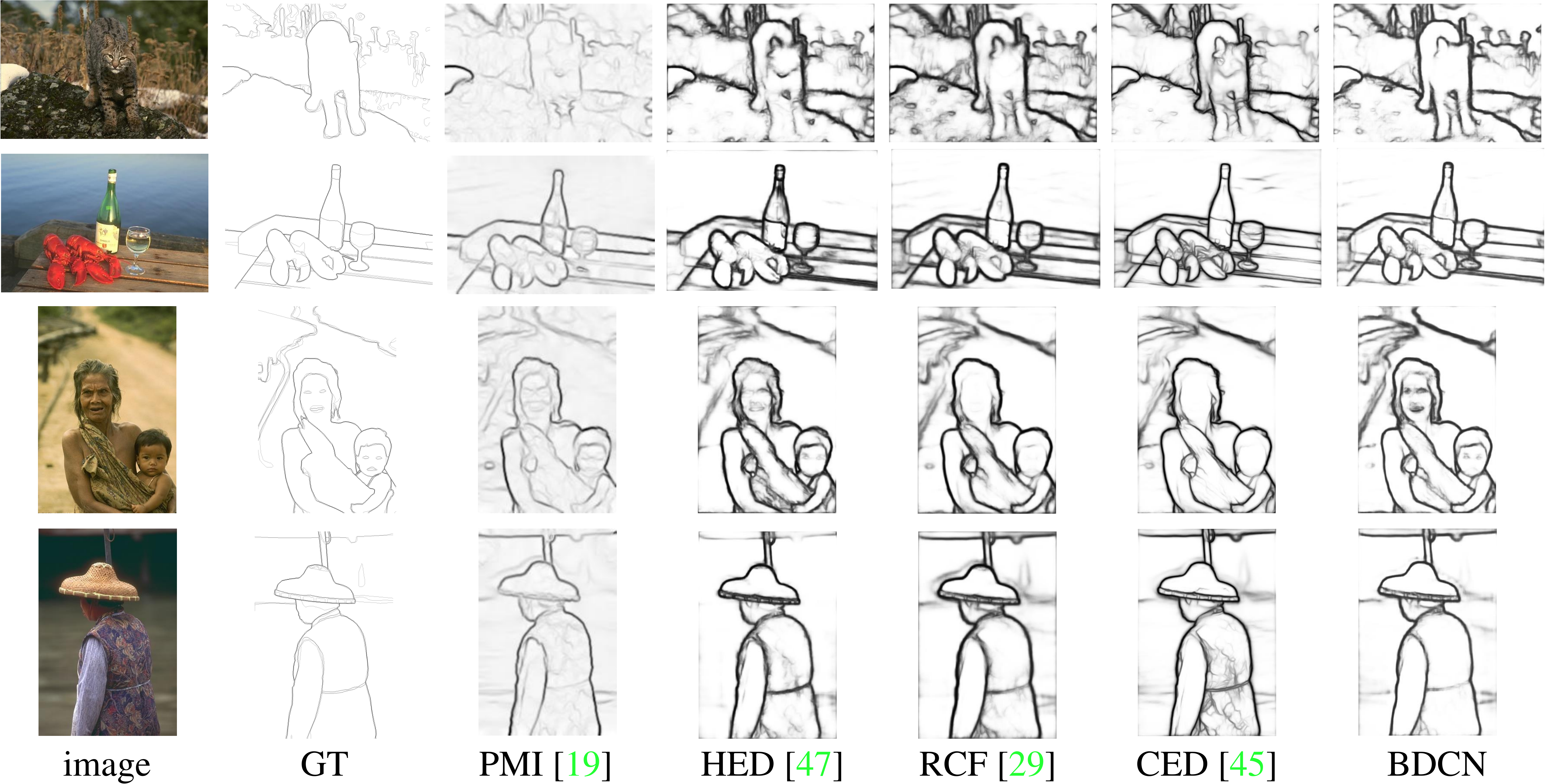,width=8.3cm}}
\caption{Comparison of edge detection results on \emph{BSDS500} test set. All the results are raw edge maps computed with a single scale input before Non-Maximum Suppression. }
\label{fig:fig_bsds}
\end{figure}

To further show the advantage of our method, we evaluate the performance of edge predictions by different intermediate layers, and show the comparison with HED~\cite{xie2015holistically} and RCF~\cite{Liu_2017_CVPR} in Table \ref{tab:side_perf}. It can be observed that, the intermediate predictions of our network also consistently outperform the ones from HED and RCF, respectively. With 5 ID Blocks, our method runs at about 22fps for edge detection, on par with most DCNN-based methods. With 4, 3 and 2 ID Blocks, it accelerates to 29 fps, 33fps, and 37fps, respectively. Fig.~\ref{fig:fig_bsds} compares some edge detection results generated by our approach and several recent ones.

\begin{table}
\begin{center}
\scriptsize
\caption{The performance (ODS) of each layer in BDCN, RCF~\cite{Liu_2017_CVPR}, and HED~\cite{xie2015holistically} on \emph{BSDS500} test set.}
\vspace{1.5mm}
\label{tab:side_perf}
\begin{tabular}{c||c c c}
\hline
Layer ID. & HED \cite{xie2015holistically} & RCF \cite{Liu_2017_CVPR} & BDCN \\
\hline
\hline
1 & 0.595 & 0.595 & \textbf{0.727} \\
2 & 0.697 & 0.710 & \textbf{0.762} \\
3 & 0.750 & 0.766 & \textbf{0.771} \\
4 & 0.748 & 0.761 & \textbf{0.802} \\
5 & 0.637 & 0.758 & \textbf{0.815} \\
\hline
fuse & 0.790 & 0.805 & \textbf{0.820} \\
\hline
\end{tabular}
\end{center}
\end{table}

\section{Conclusions}
This paper proposes a Bi-Directional Cascade Network for edge detection. By introducing a bi-directional cascade structure to enforce each layer to focus on a specific scale, BDCN trains each network layer with a layer-specific supervision. To enrich the multi-scale representations learned with a shallow network, we further introduce a Scale Enhancement Module (SEM). Our method compares favorably with over 10 edge detection methods on three datasets, achieving ODS F-measure of 0.828, 1.3\% higher than current state-of-art on \emph{BSDS500}. Our experiments also show that learning scale dedicated layers results in compact networks with a fraction of parameters, \emph{e.g.}, our approach outperforms HED~\cite{xie2015holistically} with only 1/6 of its parameters.

\section{Acknowledgement}
This work is supported in part by Peng Cheng Laboratory, in part by Beijing Natural Science Foundation under Grant No. JQ18012, in part by Natural Science Foundation of China under Grant No. 61620106009, 61572050, 91538111. We additionally thank NVIDIA for generously providing DGX-1 super-computer and support through the NVAIL program.
{\small
\bibliographystyle{ieee}
\bibliography{egbib}
}

\end{document}